\documentclass{mva_style}
\usepackage{graphicx}
\usepackage{color}
\usepackage{amsmath}
\usepackage{amssymb}
\usepackage{url}
\usepackage[colorlinks=false, pdfborder={0 0 0}]{hyperref}

\finalcopy 

\begin{document}
\title{Detection of Medial Epicondyle Avulsion in Elbow Ultrasound Images via Bone Structure Reconstruction}

\author{
Shizuka Akahori$^{1*}$ \quad
Shotaro Teruya$^{1}$ \quad
Pragyan Shrestha$^{1}$ \quad
Yuichi Yoshii$^{2}$ \\
Satoshi Iizuka$^{1}$ \quad
Akira Ikumi$^{1}$ \quad
Hiromitsu Tsuge$^{3}$ \quad
Itaru Kitahara$^{1}$ \\
$^{1}$ University of Tsukuba, Japan \quad
$^{3}$ Kikkoman General Hospital, Japan \\
$^{2}$ Tokyo Medical University, Ibaraki Medical Center, Japan \\
\texttt{akahori.shizuka@image.iit.tsukuba.ac.jp}
}


\maketitle

\section*{\centering Abstract}
\textit{
This study proposes a reconstruction-based framework for detecting medial epicondyle avulsion in elbow ultrasound images, trained exclusively on normal cases. Medial epicondyle avulsion, commonly observed in baseball players, involves bone detachment and deformity, often appearing as discontinuities in bone contour. 
Therefore, learning the structure and continuity of normal bone is essential for detecting such abnormalities.
To achieve this, we propose a masked autoencoder-based, structure-aware reconstruction framework that learns the continuity of normal bone structures. Even in the presence of avulsion, the model attempts to reconstruct the normal structure, resulting in large reconstruction errors at the avulsion site.
For evaluation, we constructed a novel dataset comprising normal and avulsion ultrasound images from 16 baseball players, with pixel-level annotations under orthopedic supervision. Our method outperformed existing approaches, achieving a pixel-wise AUC of 0.965 and an image-wise AUC of 0.967. The dataset is publicly available at: \url{https://github.com/Akahori000/Ultrasound-Medial-Epicondyle-Avulsion-Dataset}.
}

\begin{figure}[t]
\centering
  \includegraphics[width=\linewidth]{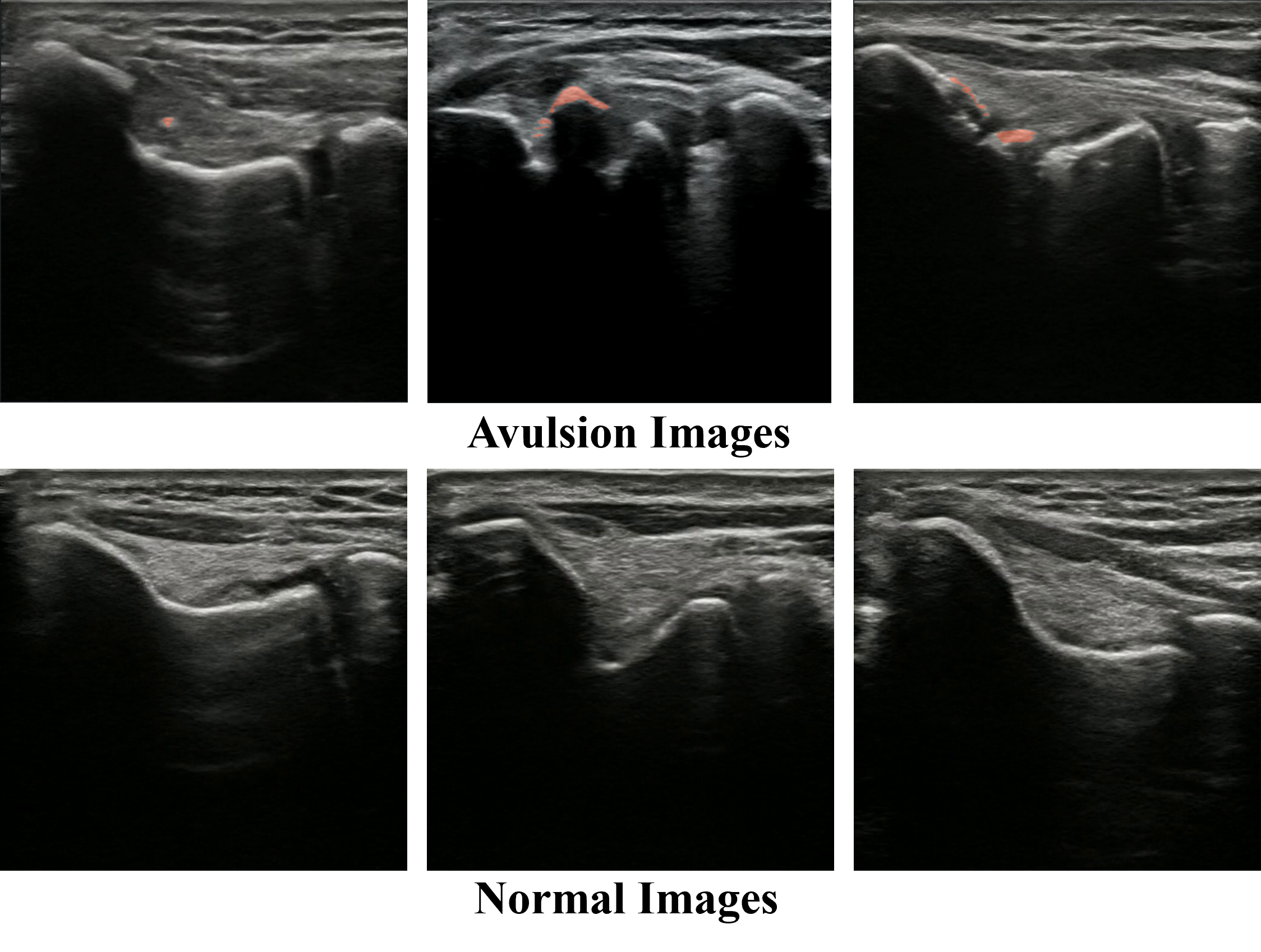}
    \caption{Ultrasound images of the medial elbow in baseball players. The upper row shows cases of medial epicondyle avulsion, while the lower row displays normal (i.e., healthy) elbows. Areas annotated in orange indicate avulsion and deformity of the humeral epicondyle. 
    Normal bone regions are illustrated on the dataset website.}
  \label{examples}
\end{figure}

\begin{figure*}[t]
  \centering
  \includegraphics[scale=0.36]{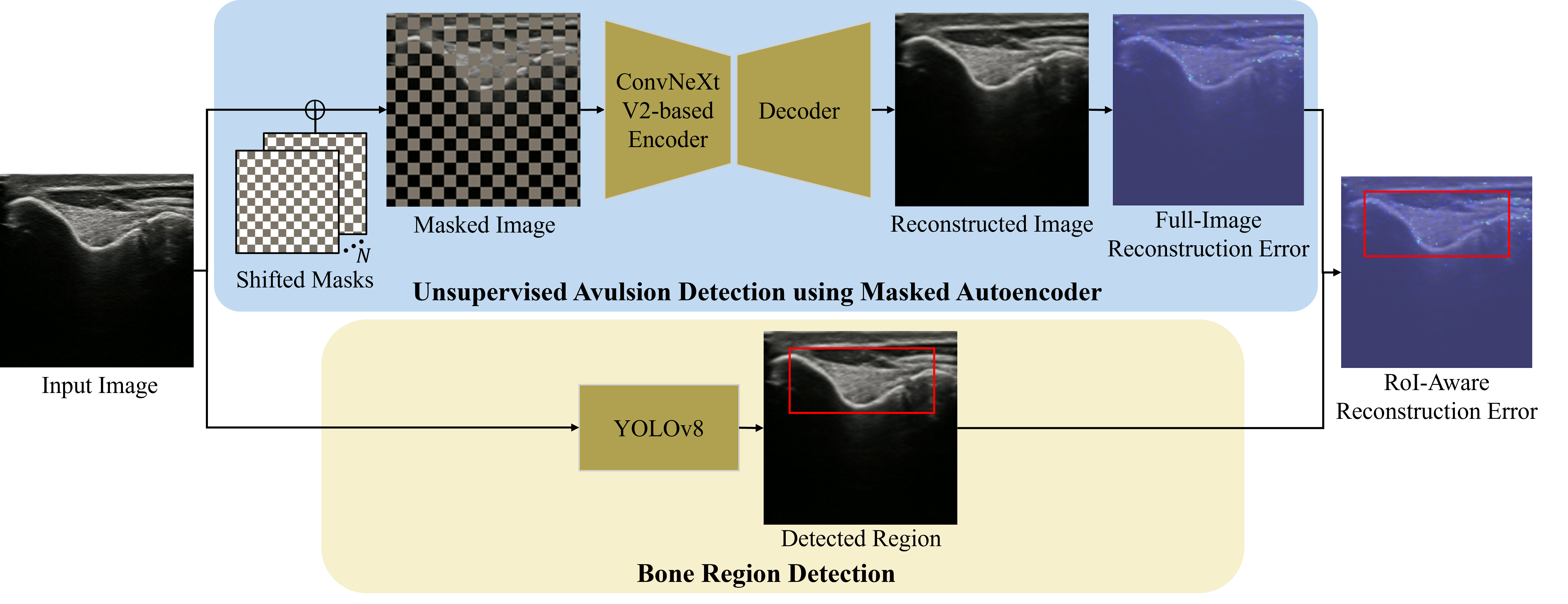}
  \caption{The proposed avulsion detection framework. The upper row illustrates a ConvNeXt V2-based masked autoencoder used to detect avulsion, while the lower row shows a bone region detection model that localizes a bounding box around the humerus and ulna. Both models are trained solely on normal samples.}
  \label{framework}
\end{figure*}

\section{Introduction}
Ultrasound imaging is widely used in sports medicine due to its non-invasive and accessible nature. In baseball players, the elbow joint is frequently examined to detect abnormalities such as avulsion fractures around the medial epicondyle of the humerus. 
Medial epicondyle avulsion is caused by acute valgus stress during throwing~\cite{shikumi} and can present a wide range of structural abnormalities, including bone deformities and loose fragments as shown in Fig.~\ref{examples}. 
Although the detection of avulsion is clinically important, it remains challenging due to morphological variability and the frequent presence of artifacts and noise in ultrasound images.
Moreover, given the limited clinical data on avulsion cases, it is necessary to develop the detection model solely on normal images.

Several unsupervised anomaly detection methods have been applied to medical imaging~\cite{ae_med, vitmae_med, patchcore_med, gan_med, cflowad-med}. 
These include normalizing flow-based methods such as CFLOW-AD~\cite{cflowad}; reconstruction-based methods such as autoencoders~\cite{ae}; memory bank-based methods such as PatchCore~\cite{patchcore}; and anomaly synthesis-based methods such as SimpleNet~\cite{simplenet}.
For ultrasound imaging, generative adversarial network~\cite{gan_us}, autoencoder~\cite{swae_us}, and diffusion~\cite{diffusion_us} have been utilized.
While diffusion-based methods have recently shown promising results, they typically require substantially more data and computational resources, limiting their applicability in clinical settings where such resources are often limited.


In elbow ultrasound imaging, several studies have focused on detecting osteochondritis dissecans on the lateral side~\cite{ultrasound-elbow1, ultrasound-elbow2, ultrasound-elbow3, ultrasound-elbow4}. 
However, research on the medial elbow is limited, and there is no publicly available dataset to detect humeral epicondyle avulsion.


To address these gaps, we propose a framework for detecting medial epicondyle avulsion in elbow ultrasound images, trained solely on normal cases. 
The main contributions are follows.
\begin{itemize}
\item We propose a ConvNeXtV2-based masked autoencoder that reconstructs both masked and unmasked regions from surrounding areas to learn the continuous anatomical structure of bone. A shifted chessboard masking strategy is integrated to enhance local diversity.
\item We incorporate a bone region detection module trained on normal images to compute the reconstruction error specifically within the region.
\item For evaluation, we introduce a medial epicondyle avulsion dataset comprising normal and avulsion ultrasound images of baseball players, with pixel-level annotations under orthopedic supervision.
\end{itemize}




\section{Related Work}
Masked Autoencoder (MAE) is a self-supervised learning method that learns visual representations from unlabeled data. The original MAE~\cite{vit_mae}, which uses a Vision Transformer (ViT)~\cite{vit} and is referred to as ViT MAE, randomly masks a large portion of the input image and reconstructs the masked regions.
To adapt MAE to Convolutional Neural Networks (CNNs)~\cite{cnn}, ConvNeXt V2~\cite{convnextv2} has been proposed. ConvNeXt V2 employs sparse convolutions that operate only on visible regions, skipping masked areas. A Fully Convolutional Masked Autoencoder (FCMAE) with a CNN decoder is used to reconstruct the images. Its CNN-based architecture enables efficient computation while effectively capturing local spatial features.

\section{Method}
We propose a framework for detecting medial epicondyle avulsion in elbow ultrasound images.
Fig.~\ref{framework} illustrates our approach: the upper row represents an unsupervised anomaly detection pipeline using a ConvNeXt V2-based masked autoencoder, while the lower row depicts a semi-supervised object detection pipeline using YOLOv8~\cite{yolov8}. Both models are trained exclusively on normal images and aim to detect avulsion-related bone abnormalities by measuring the reconstruction error within the region of interest (RoI) identified by the object detection model.
This region-specific reconstruction error not only focuses the analysis on areas where bone avulsion and deformity are likely to occur, but also helps reduce the impact of noise and artifacts commonly found in ultrasound images.

\subsection{ConvNeXt V2-based Masked Autoencoder}
Our masked autoencoder uses a ConvNeXt V2-based encoder and a linear decoder.
Our model differs from the original ConvNeXt V2 in three key ways.
First, while the original reconstructs only the masked regions and computes loss accordingly, our model reconstructs both masked and unmasked regions, computing the loss over the entire image to encourage smoother reconstruction and learn normal bone continuity.
Second, our model dynamically shifts the masking positions instead of masking at fixed patch locations.
Third, our model uses dense convolution instead of sparse convolution.
The process is outlined below. 

To begin with, an input image is masked using a chessboard pattern with randomly shifted mask positions at the pixel. 

\begin{equation}
X' = M \odot X,
\end{equation}
where \( X,  X' \in \mathbb{R}^{C \times H \times W} \) are the input and masked images, and \( M \in \{0,1\}^{H \times W} \) is a chessboard mask randomly selected from a set of $N$ predefined masks, with \( C \), \( H \), and \( W \) denoting the number of channels, height, and width, respectively.
The masked regions of $X'$ are filled with gray values.

Next, the masked image \( X' \) is processed by the encoder \( E \) and a 1D linear decoder \( D \), yielding the reconstructed output $R$:

\begin{equation}
R = D(E(X')).
\end{equation}

Then, the loss $L$ across the entire image is computed with the Mean Squared Error.

\begin{equation}
L = \frac{1}{C \cdot H \cdot W} \sum_{c,h,w} (R_{c,h,w} - X_{c,h,w})^2.
\end{equation}


\subsection{Bone Region Detection}
YOLOv8 model is trained to obtain a bounding box \( B \subset \mathbb{R}^2 \) that surrounds the epicondyle and the ulna in the input image, along with its confidence score \( S \).

\subsection{Reconstruction Error}
During inference, to ensure that the entire image, including abnormal regions, is masked at least once, multiple reconstructed images are generated by shifting the chessboard mask, and the reconstruction errors are averaged.
The full-image pixel-wise reconstruction error $E_{\text{pixel}}^{F}$ is defined as:
\begin{equation}
E_{\text{pixel}}^{F}(h, w) = \frac{1}{N^{\prime}} \sum_{i=1}^{N^{\prime}} E_{i,\text{pixel}}(h, w),
\end{equation}
where, $N^{\prime}$ denoting the number of the shifted masks, and the i-th reconstruction error with the i-th mask is $E_{i,\text{pixel}}(h, w) = (R_{i}(h, w) - X(h, w))^2$. 


Focusing on regions relevant to avulsion, we define the reconstruction error within the detected bone region as the RoI-aware reconstruction error $E_{pixel}^{R}$.
When the confidence score \( S \) exceeds or equals a confidence score threshold \( \tau \), the error within the bounding box \( B \) is retained.
\begin{equation}
E_{pixel}^{R}(h, w) =
\begin{cases}
E_{pixel}^F(h, w) & \text{if } (h, w) \in B \\
0 & \text{otherwise}
\end{cases}
\end{equation}
When \( S < \tau \), \( E_{pixel}^{R} \) is set to the full-image error \( E_{pixel}^{F} \).

The full-image and RoI-aware image-wise reconstruction errors $E_{image}^{F}$, $E_{image}^{R}$ are computed by averaging the top \( k\%\) of of the highest pixel values in \( E_{pixel}^{F} \) and \( E_{pixel}^{R} \), respectively.

\section{Avulsion Dataset Construction}\label{dataset_const}
We constructed an ultrasound dataset to detect medial epicondyle avulsion from 16 current and former baseball players (8 normal, 8 avulsion), aged 18–36, to evaluate the method’s ability to identify subtle elbow abnormalities in individuals with similar physical characteristics.
Each participant contributed 14 images, totaling 112 normal and 112 avulsion images.
The images were acquired using a musculoskeletal ultrasound system with an 11-MHz linear probe, operated by three experienced orthopedic surgeons. During the examination, each participant lay supine with the elbow flexed at 90 degrees. A surgeon placed the gel-applied probe along the direction from the medial epicondyle of the humerus to the ulnar collateral ligament, which captures commonly examined areas for diagnosing medial elbow pathologies~\cite{michinobu}.
All images were annotated at the pixel level under expert orthopedic supervision.
The examples are shown in Fig.~\ref{examples}. 


\section{Experiment}\label{experiment}
The backbone of the masked autoencoder was ConvNeXt V2 nano~\cite{timm}.
The input image was resized to $128\times128$ pixels, and each square in the chessboard masks was \( 8 \times 8 \) pixels. 
The mask was shifted by 1 pixel with $N = 128$ for training, and by 2 pixels with $N^{\prime} = 32$ for testing.
Data augmentation included random rotation, scaling, cropping, shift, mean padding, reflection padding, and enlargement. 
The confidence score threshold $\tau$ was set to 0.5.
The masked autoencoder was trained on 2,997 images and validated on 1,204 normal images from the medial elbow dataset~\cite{shizuka}, and the bone detection model was trained on the same data using bounding boxes derived from the landmark annotation~\cite{shizuka}.
The image-wise reconstruction error was calculated with $k=0.3\%$.
Pixel-wise and image-wise Area Under the ROC Curve (AUC) scores were used as evaluation metrics. 



\begin{figure*}[t]
  \centering
  \includegraphics[scale=0.8]{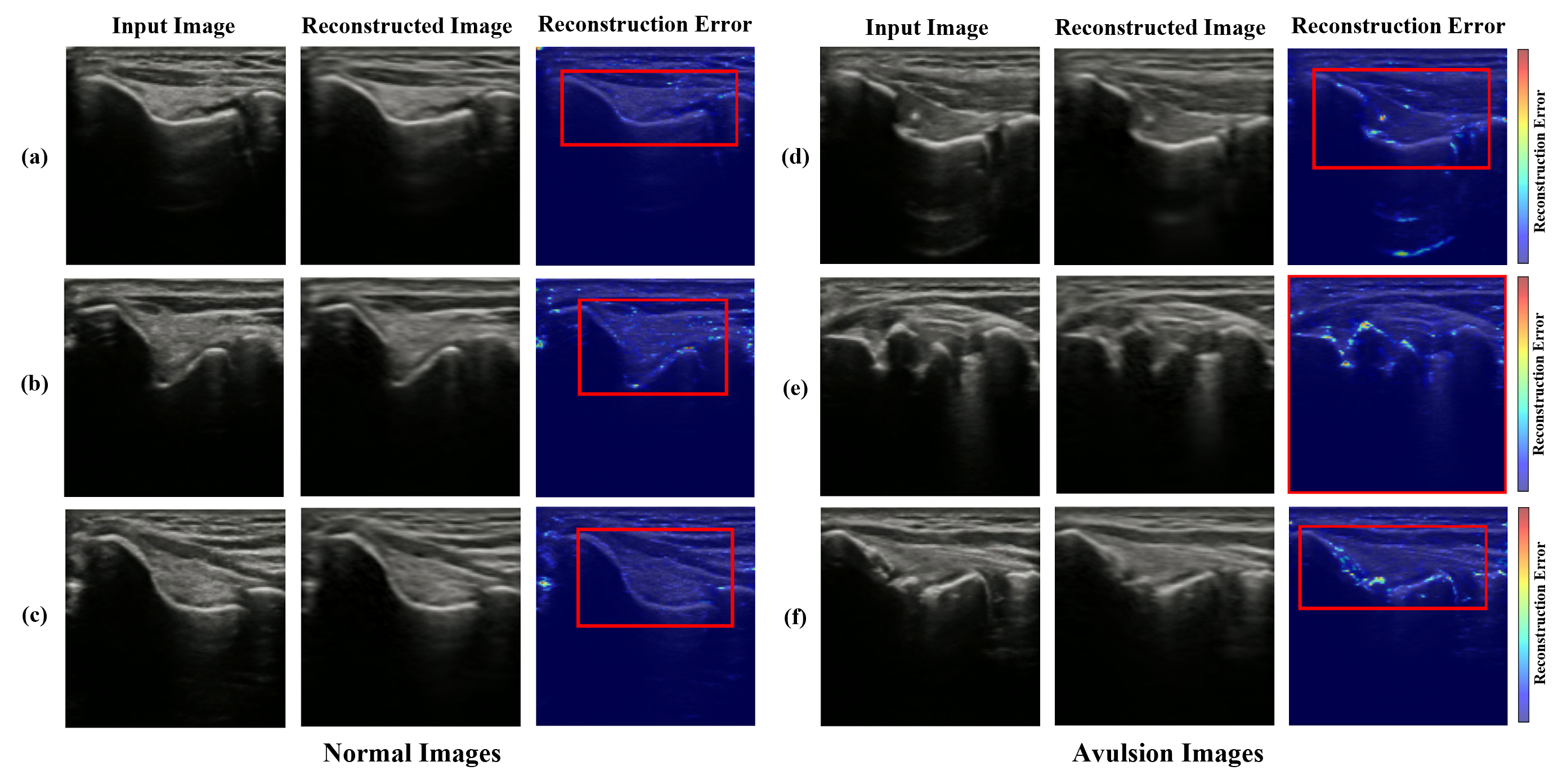}
  \caption{Visualization of our results. (a)–(c) show normal samples, and (d)–(f) show avulsion cases. Each sample includes: an input image (left), a reconstructed image from a masked input using a single chessboard mask (middle), and an averaged reconstruction error map from shifted masks overlaid on the input (right). Detected bone regions are shown in red boxes. Ground truth avulsion annotations are provided in Fig.~\ref{examples}.}
  \label{results}
\end{figure*}

\section{Results}\label{results}
Pixel-wise and image-wise AUC scores measured under the Full-image and RoI-aware settings are shown in Tables~\ref{table1} and~\ref{table2}.
Our method achieved the highest pixel-wise AUC in both settings, demonstrating superior localization of abnormal regions relevant to avulsion.
For image-wise AUC, although the Full-image score was relatively lower, our method outperformed the others in the RoI-aware setting.  
Fig.~\ref{results} shows reconstructed images and error maps, indicating that the images are well reconstructed despite half of the input pixels being masked. In addition, avulsed bone regions exhibit higher error values than normal bone regions.  
Notably, the relatively high error values outside the bone regions in Fig.~\ref{results}(b), (c), and (d), mostly caused by artifacts, are excluded from the RoI-aware error calculation, contributing to improved performance.

\begin{table}[t]
  \caption{Pixel-wise AUC scores for avulsion detection under Full-image and RoI-aware settings.}
  \begin{center}
    \begin{tabular}{c | c c }
      \hline
      \hline
      {Method} & {$AUC_{Full}\uparrow$} & 
      {$AUC_{RoI}\uparrow$} \\
      \hline
        Patchcore         & 0.947 & 0.956 \\ 
        CFLOW-AD         & 0.912 &  0.934 \\ 
        SimpleNet         & 0.859 &  0.852 \\ 
        ViT MAE          & 0.842 & 0.890 \\
        ConvNeXt V2 FCMAE  & 0.206 & 0.635 \\
        Ours                & \textbf{0.951} & \textbf{0.965} \\
      \hline
      \hline
    \end{tabular}
    \label{table1}
  \end{center}
\end{table}

\begin{table}[t]
  \caption{Image-wise $AUC$ scores for avulsion detection under Full-image and RoI-aware settings.}
  \begin{center}
    \begin{tabular}{c | c c }
      \hline
      \hline
      {Method} & {$AUC_{Full}\uparrow$} & 
      {$AUC_{RoI}\uparrow$} \\
        \hline
        Patchcore         & \textbf{0.931} & 0.954 \\ 
        CFLOW-AD          & 0.817 & 0.834 \\   
        SimpleNet         & 0.822  & 0.836 \\
        ViT MAE          & 0.827 & 0.898 \\
        ConvNeXt V2 FCMAE  & 0.409 & 0.549 \\
        Ours                & 0.804 & \textbf{0.967} \\
      \hline
      \hline
    \end{tabular}
    \label{table2}
  \end{center}
\end{table}



\section{Discussion}\label{Discussion}
First, we evaluated the effectiveness of the masking strategy by training an autoencoder, i.e., our network without masking. The resulting RoI-aware AUC scores were 0.826 (pixel-wise) and 0.748 (image-wise), both lower than those achieved with masking. This suggests that masking helps the model learn anatomical structures from neighboring patches and enhances avulsion detection. 
Finer shifts of the chessboard mask led to better results, suggesting that the shifting strategy improves local diversity. Inference time using $N^{\prime} = 32$ was 0.0142 s/image on an RTX 3090.
Second, in the image-wise evaluation of the proposed method, the full-image AUC was relatively low, while the ROI-aware AUC improved significantly. This is because artifacts often observed outside the bone region in normal test data (e.g., Fig.~\ref{results}(c)) were excluded in the ROI-aware error computation.
Third, while PatchCore showed the second-best performance, its anomaly maps exhibited widespread high scores, including in non-bone regions, possibly due to its patch feature-based scoring.
Future work could include avulsion detection in pediatric elbows and varying chessboard mask sizes.





\section{Conclusion}\label{Conclusion}
We presented a reconstruction-based framework for detecting medial epicondyle avulsion in elbow ultrasound, trained exclusively on normal images. The proposed method combines a ConvNeXtV2-based masked autoencoder with a bone localization module that computes region-specific reconstruction errors.  
Experiments on a clinically annotated dataset demonstrated superior performance in both pixel-wise and image-wise detection.   
These results highlight the potential of modeling bone continuity for detecting avulsions without requiring abnormal training samples.

\section{Acknowledgements}
The authors would like to thank Dr. Ryuhei Michinobu for his support in data collection and for providing valuable medical expertise throughout this study.  
We are also grateful to Dr. Yasukazu Totoki for his overall guidance and financial support.
This work was supported by JST SPRING, Grant Number JPMJSP2124, JSPS KAKENHI Grant Number JP24K14397, and the Multidisciplinary Cooperative Research Program in CCS, University of Tsukuba.

\end{document}